\documentclass{article}
\usepackage{amssymb}
\usepackage{graphicx}

\usepackage[final]{corl_2025} 
\PassOptionsToPackage{sort&compress}{natbib}
\usepackage{wrapfig}

\usepackage{enumitem}
\usepackage{booktabs}
\usepackage{soul}
\usepackage{xcolor}
\usepackage[ruled,vlined,linesnumbered]{algorithm2e}
\usepackage{wrapfig}
\usepackage{graphicx}
\usepackage[font=footnotesize]{subcaption}
\usepackage[font=footnotesize]{caption}
\usepackage{amsmath,amssymb}
\usepackage{algpseudocode}

\usepackage{xcolor}
\usepackage{soul}  
\usepackage{color, soul}
\usepackage[customcolors]{hf-tikz}
\usepackage{dsfont} 
\usepackage{colortbl}
\usepackage{multirow}
\usepackage{lipsum}
\DeclareSymbolFont{rsfs}{U}{rsfs}{m}{n}
\DeclareSymbolFontAlphabet{\mathscrsfs}{rsfs}

\definecolor{perfblue}{RGB}{64, 114, 175}
\usepackage{transparent}
\usepackage{wrapfig}

\newcommand{\methodname}{{\textsc{X-Sim}}}

\title{\methodname: Cross-Embodiment \\Learning via Real-to-Sim-to-Real}

\author{
    \begin{tabular}{c}
        Prithwish Dan\thanks{Equal Contribution} \quad
        Kushal Kedia\footnotemark[1] \quad
        Angela Chao \quad
        Edward W. Duan \quad \\
        Maximus A. Pace \quad
        Wei-Chiu Ma \quad
        Sanjiban Choudhury \\
        \normalfont{Cornell University}
    \end{tabular}
}

\begin{document}
\maketitle



\begin{abstract}
Human videos offer a scalable way to train robot manipulation policies, but lack the action labels needed by standard imitation learning algorithms. Existing cross-embodiment approaches try to map human motion to robot actions, but often fail when the embodiments differ significantly. We propose \methodname, a real-to-sim-to-real framework that uses object motion as a dense and transferable signal for learning robot policies. \methodname\ starts by reconstructing a photorealistic simulation from an RGBD human video and tracking object trajectories to define object-centric rewards. These rewards are used to train a reinforcement learning (RL) policy in simulation. The learned policy is then distilled into an image-conditioned diffusion policy using synthetic rollouts rendered with varied viewpoints and lighting. To transfer to the real world, \methodname\ introduces an online domain adaptation technique that aligns real and simulated observations during deployment. Importantly, \methodname\ does not require any robot teleoperation data. We evaluate it across 5 manipulation tasks in 2 environments and show that it: (1) improves task progress by 30\% on average over hand-tracking and sim-to-real baselines, (2) matches behavior cloning with 10$\times$ less data collection time, and (3) generalizes to new camera viewpoints and test-time changes. 
Code and videos are available at \href{https://portal-cornell.github.io/X-Sim/}{https://portal-cornell.github.io/X-Sim/}.

\end{abstract}

\section{Introduction}

Human videos offer a natural and scalable source of demonstrations for robot policy learning.
However, recent advances in robot foundation models~\cite{Kim2024OpenVLAAO, Intelligence2025pi} rely entirely on large-scale datasets of robot embodiments~\cite{padalkar2023open, khazatsky2024droid}. Collecting such data requires labor-intensive and expensive teleoperation to provide high-quality expert demonstrations, making it intractable to scale across diverse tasks and environments. In contrast, human videos (e.g. from YouTube) are abundant and capture a wide range of tasks in natural environments.

Despite their potential, human videos cannot be directly used in widely-adopted imitation learning pipelines~\cite{chi2023diffusion, Zhao2023LearningFB}, as they lack explicit robot action labels. To bridge this gap, prior work attempts to map human trajectories to robot actions, typically assuming visual or kinematic compatibility.
Some methods retarget human hand motion to the robot's end-effector~\cite{Bharadhwaj2023ZeroShotRM}, but this assumes that human movements are feasible for the robot to replicate~\cite{Ren2025MotionTA}, which is rarely the case in practice. Other methods reduce the human-robot visual gap by overlaying robot arms on human videos~\cite{Lepert2025SHADOWLS, Lepert2025PhantomTR}, but these rely on solving inverse kinematics, which may be ill-posed due to embodiment mismatch.
Another line of work directly translates human videos into robot actions~\cite{Jang2022BCZZT, Jain2024Vid2RobotEV, wang2023mimicplay}, but requires paired human-robot demonstrations, which are expensive and difficult to collect at scale.

We tackle the problem of generating robot training data from action-less human videos. \textbf{\emph{Our key insight is that, while human actions are unavailable, the object motion they produce provides a dense supervisory signal for training robot policies in simulation.}}
By reconstructing a photorealistic simulation~\cite{Huang20242DGS} of the human video and tracking object trajectories~\cite{Wen2023FoundationPoseU6}, we define object-centric reward functions that guide RL agents to reproduce the effects of human behavior — even when the robot must take entirely different actions. This enables distillation into real-world image-conditioned robot policies \textbf{\emph{without any robot teleoperation data}}.

\begin{figure}[t]  
  \centering
  \includegraphics[width=0.99\textwidth]{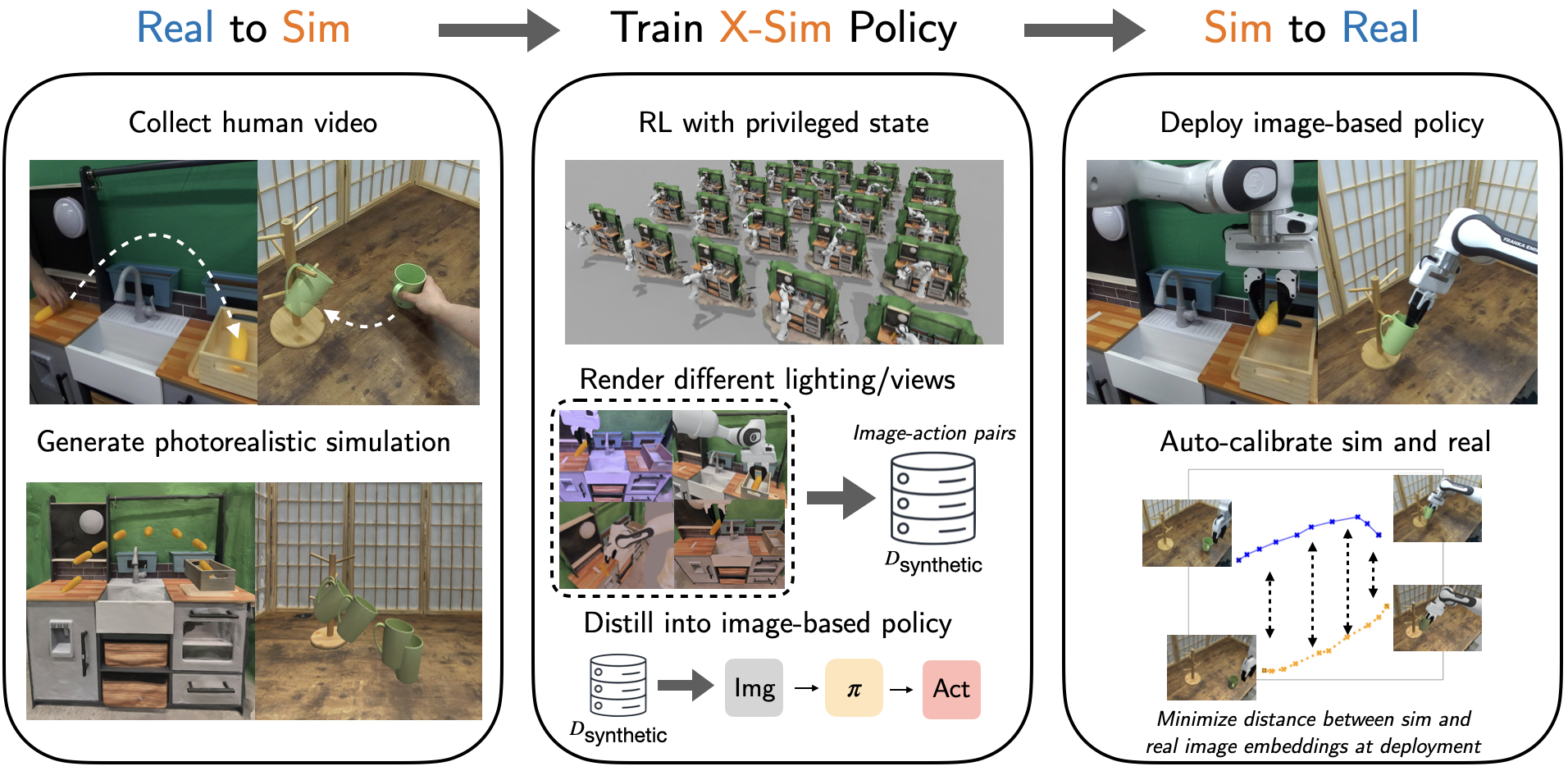}
  \caption{\textbf{Overview of \methodname:} We introduce \methodname, a real-to-sim-to-real framework that bridges the human-robot embodiment gap by learning robot policies. \textbf{Real-to-Sim}. We generate photorealistic simulation using object-centric rewards generated from human videos. \textbf{Training X-Sim}. We first train RL policies with privileged state using GPU-parallelized environment. Then, we collect a diverse image-action dataset use it to distill behaviors into an image-conditioned policy. \textbf{Sim-to-Real.} Image-based policy is deployed in the real-world. Its observation encoder automatically calibrates itself by aligning real and sim image observations at test-time.}
  \label{fig:intro}
  \vspace{-3mm}
\end{figure}

We propose \methodname, a real-to-sim-to-real framework that bridges the human-robot embodiment gap by learning robot policies in simulation on rewards generated from human videos (Fig.~\ref{fig:intro}). \methodname\ first extracts object states from a RGBD human video and transfers them into a photorealistic simulation. It defines a dense object-centric reward to efficiently train state-based RL policies in simulation. \methodname\ generates a large synthetic dataset of paired image-action data by rolling out the trained RL policy and rendering the resulting scenes under varied robot poses, object states, viewpoints, and lighting.  Using this dataset, it trains an image-conditioned diffusion policy and transfers directly to the real-world without needing any real robot action data. To narrow the sim-to-real gap at deployment, \methodname\ utilizes an online domain adaptation technique to align the robot's real world and simulation observations. Our contributions are summarized as follows:
\vspace{-2mm}
\begin{enumerate}[leftmargin=*]
\itemsep0em 
\item We propose \methodname, a real-to-sim-to-real framework that learns image-based robot policies from action-less human videos by tracking object states and matching their motion in simulation.
\item We introduce an online domain adaptation technique to continually reduce the sim-to-real gap by aligning real-world observations with simulation at test time, enabling robust sim-to-real transfer. 
\item We evaluate \methodname\ across 5 manipulation tasks in 2 environments, showing that it (1) improves task progress by 30\% on average over hand-tracking and sim-to-real baselines, (2) matches behavior cloning with 10x less data collection time, and (3) enables generalization to test-time environment changes, including novel camera viewpoints.
\end{enumerate}

\vspace{-3mm}
\section{Related Work}
\vspace{-2mm}
\textbf{Imitation Learning.} 
Imitation learning, particularly behavior cloning (BC), is the dominant paradigm for training visuomotor robot policies. Recent algorithms like Diffusion Policy~\cite{chi2023diffusion} and ACT~\cite{Zhao2023LearningFB} achieve state-of-the-art results by learning from expert demonstrations consisting of image-action pairs. However, these methods typically require collecting data via human teleoperation of the specific target robot, using kinesthetic teaching~\cite{Billard2006DiscriminativeAA}, wearable devices~\cite{Iyer2024OPENTA}, or specialized control interfaces~\cite{Levine2015EndtoEndTO, Wu2023GELLOAG, Zhao2024ALOHAUA}. Recent efforts have attempted to build large robotic dataset across different robot embodiments~\cite{padalkar2023open, khazatsky2024droid} leading to the development of foundation models~\cite{Kim2024OpenVLAAO, Intelligence2025pi} for robotic control. Still, scaling up such datasets remains a significant challenge because of the heavy reliance on robot teleoperation.  While UMI~\cite{Chi2024UniversalMI} proposes hand-held grippers to collect data without direct robot involvement, these demonstrations can be dynamically infeasible for robots and still require active collection in lab settings. In contrast, our approach bypasses the need for robot action data entirely by leveraging human videos to generate synthetic robot data.

\textbf{Learning from Human Videos.} The ease of collecting human videos has motivated interest in learning robot motion directly from them. Common strategies include retargeting hand motion~\cite{telekinesis,Shaw2022VideoDexLD, arunachalam2022dime, Ye2022LearningCG, Bharadhwaj2023ZeroShotRM}, reducing the visual gap via inpainting~\cite{bahl2022human, Lepert2025SHADOWLS, Lepert2025PhantomTR}, or using pretrained open-world vision models for constructing object-relative hand trajectories~\cite{zhu2024vision, Vitiello2023OneShotIL, Li2024OKAMITH}. All of these methods rely on the robot's capability to match its end-effector with the human's hand positions, which often falls down in practice due to large embodiment differences. Hierarchical frameworks~\cite{wang2023mimicplay, bharadhwaj2024track2act, Bharadhwaj2023TowardsGZ} learn high-level plans instead, while one-shot imitation methods~\cite{Jang2022BCZZT, Jain2024Vid2RobotEV,Bharadhwaj2024Gen2ActHV} learn from prompt videos. These methods typically require human-robot paired data or self-supervised alignment from unpaired data~\cite{kedia2024one, xu2023xskill}. In either case, a common limitation among these methods is the need for robot teleoperation data to guide low-level control~\cite{Ren2025MotionTA}. RL provides an alternative, using video~\cite{Zakka2021XIRLCI} similarity, language matching~\cite{Shao2020Concept2RobotLM} or object tracking~\cite{Patel2022LearningTI} for rewards, but suffers from the sim-to-real gap. Cross-embodiment RL~\cite{kumar2022inverse, Gzey2024BridgingTH} methods that have been deployed on real robots require object tracking at test-time which can be brittle to noisy observations. Instead, we leverage a real-to-sim-to-real pipeline to directly transfer image-based policies from simulation.

\textbf{Real-to-Sim-to-Real.} Advances in 3D computer vision have enabled the development of photorealistic, physically accurate simulations from real-world data. Recent works increasingly use real-to-sim methods to learn robot behaviors in simulation. For instance, RialTo~\cite{Torn2024ReconcilingRT} trains RL policies in simulation to improve policy robustness, using point cloud inputs for real-world deployment. ResiP~\cite{Ankile2024FromIT} learns residual actions in simulation starting from an image-based policy trained in the real world. However, both these approaches still require real-world robot data collection. To directly learn actions, motion planners are used in simulation but deployed open-loop in the real world~\cite{Patel2025ARA, Kerr2024RobotSR}. More recently, real-to-sim-to-real has been applied to learn from human videos~\cite{Ye2025Video2PolicySU, Ga2025CrossingTH}. However, Video2Policy~\cite{Ye2025Video2PolicySU} only extracts the initial and final object states from human videos, and relies on object segmentation masks at test time for policy transfer. Human2Sim2Robot~\cite{Ga2025CrossingTH} defines rewards for RL using object state tracking from videos, but does not use a photo-realistic simulation. However, real-world deployment additionally requires object tracking at test time. RL training also requires tracking human hand trajectories for guiding the policy, and is applied only to dexterous hands with minimal embodiment gap. Our work offers distinct advantages over these methods: (a) we bypass the need for robot teleoperation data and human hand tracking for RL training, and (b) we transfer image-based policies from simulation to the real world using environment randomization and domain adaptation methods.

\vspace{-3mm}
\section{Approach}
\vspace{-1mm}
\label{sec:approach}

\methodname\ addresses the problem of training real-world, image-conditioned robot policies from action-less RGBD human videos by using object motion as a transferable supervision signal. The framework consists of three stages: (1) reconstructing a photorealistic simulation environment from the human video and extracting object trajectories to define dense object-centric rewards; (2) training privileged-state reinforcement learning (RL) policies in simulation to reproduce the observed object motion, and generating synthetic image-action data through rollouts; and (3) distilling the learned behaviors into an image-conditioned diffusion policy, and deploying it in the real world with an online domain adaptation technique that continually aligns simulated and real observations.

\subsection{Real-to-Sim Transfer from Human Videos}

\begin{figure}[t]  
  \centering
  \includegraphics[width=0.99\textwidth]{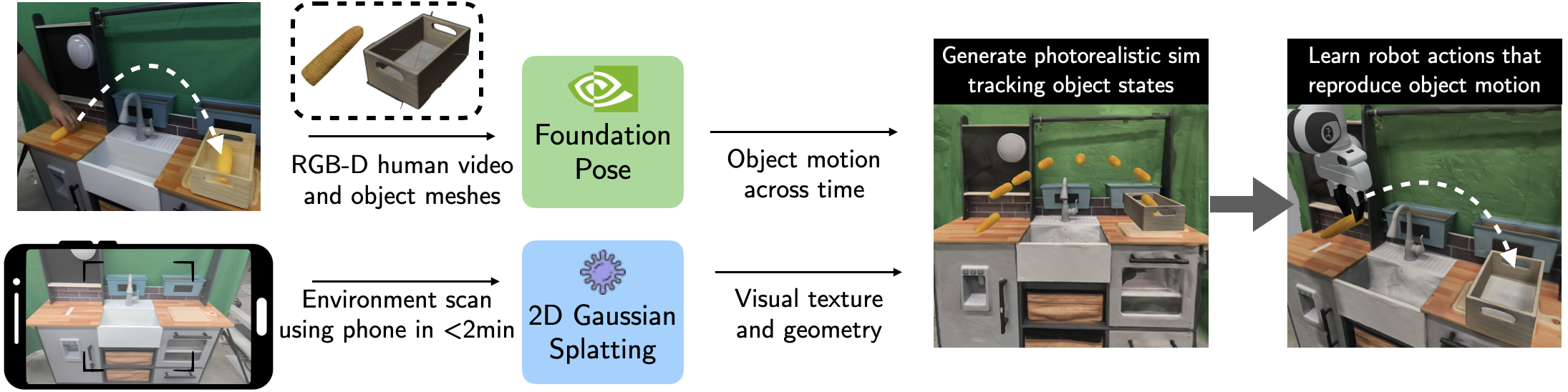}
  \caption{\textbf{Real-to-Sim:} \methodname\ reconstructs a photorealistic environment with accurate geometry from multi-view images. It tracks object motion across time from an RGBD human video to define a dense object-centric reward function to train RL policies in simulation.}
  \label{fig:approach-real2sim}
\end{figure}

While human videos do not provide direct supervision for robot actions, the resulting object motion can serve as a transferable task specification. 
This stage reconstructs a realistic simulation environment and extracts object trajectories from the human video, enabling policy learning through object-centric rewards (Fig.~\ref{fig:approach-real2sim}).

\textbf{Object Pose Tracking.} We first use an off-the-shelf 3D scanning app~\cite{polycam2020} to obtain high-fidelity object meshes of items being manipulated. To track these objects across the video, we employ FoundationPose~\cite{Wen2023FoundationPoseU6} which takes in the human video, the 3D mesh of each object and a 2D mask identifying the objects in the first frame of the video, generated by Segment Anything (SAM)~\cite{Kirillov2023SegmentA}. FoundationPose tracks the position and rotation of each object over the course of the video. Formally, given a human video $\mathbf{v_H} = \{v^t_H\}_{t=1}^T$ where $v^t_H \in \mathbb{R}^{H \times W \times 4}$ is the RGBD image at timestep $t$, we convert $\mathbf{v_H}$ into $\mathbf{s_H} = \{s^t_H\}_{t=1}^T$ where $s^t_H \in {\rm{SE}}(3)^K$ represents the position and rotation of $K$ objects being tracked in the scene.

\textbf{Environment Reconstruction.} We next construct a geometrically accurate and photorealistic environment mesh using 2D Gaussian Splatting~\cite{Huang20242DGS}, an open-source module that performs 3D reconstruction from multi-view images. This reconstructed environment is then transferred directly into the ManiSkill~\cite{Mu2021ManiSkillGM} simulator along with the object states. Physical properties and dynamics are set to default values for simplicity, though the approach is compatible with system identification~\cite{lim2021planar, chang2020sim2real2sim}, domain randomization~\cite{chebotar2019closing, peng2018sim, tobin2017domain}, and additional methods to handle articulations~\cite{hsu2023ditto, xia2025drawer}.

\begin{figure}[t]  
  \centering
  \includegraphics[width=0.99\textwidth]{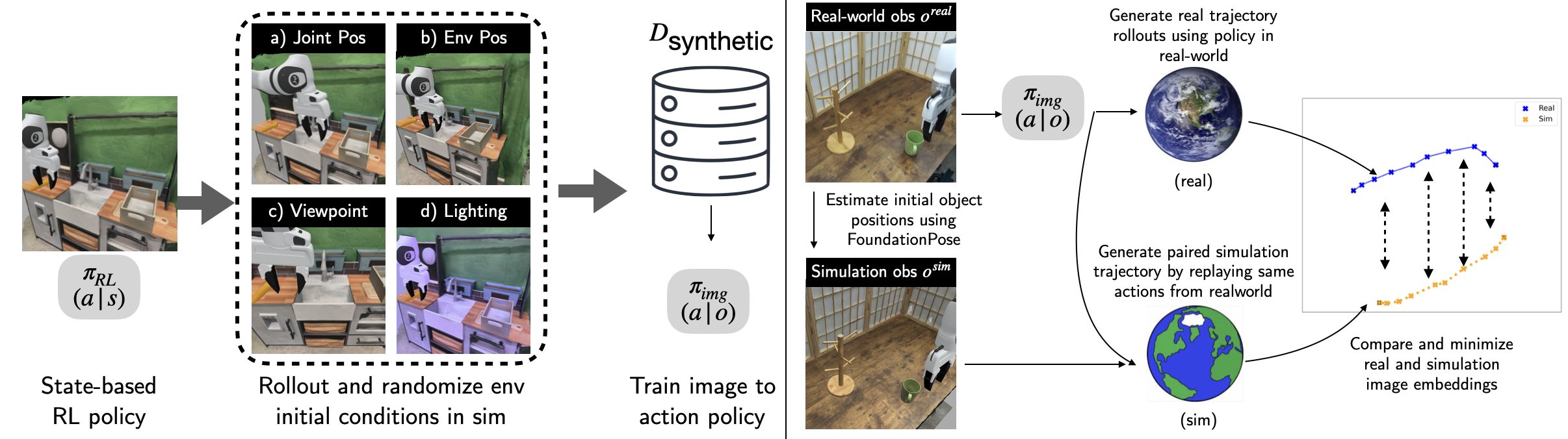}
  \caption{\textbf{Sim-to-Real:} \textbf{(Left)} \methodname\ distills privileged-state policies into image-conditioned policies by generating and a synthetic image-action dataset using multiple environment randomizations. \textbf{(Right)} During deployment, real policy rollouts are replayed in simulation to generate paired images across real and sim. Their discrepancy is utilized to minimize and calibrate the sim-to-real visual gap.}
  \label{fig:approach-sim2real}
\end{figure}
\subsection{Generating Robot Actions in Simulation}
To bridge the embodiment gap and obtain robot actions to complete the task specified by the human video, we define an object-centric reward function to train a privileged-state policy via RL. Then, we rollout the policy and render the scene under varied robot poses, object states, viewpoints, and lighting conditions to collect a synthetic dataset of image-action pairs (Fig.~\ref{fig:approach-sim2real}, Left).

\textbf{Defining Object-Centric Rewards.} We use the human video's object pose trajectory to define an object-centric reward function, with each goal indicating where the objects should be positioned and oriented. The reward encourages the robot to move objects toward their next desired pose, as specified by the human trajectory. For a goal state $s_H^B$, the reward is defined as:
\begin{equation}
r_{\rm goal} \propto - d_{pos}(s^{B}_H,s^{t}_R) - d_{rot}(s^{B}_H, s^{t}_R)
\end{equation}
where the objects current positions and rotations are encouraged to match the next goal. In practice, there is an additional default reward which brings the robot end-effector near the relevant objects $r_{\rm obj} = r_{\rm approach} + r_{\rm goal}$. The policy's privileged state includes information about the current object states, goal states, and robot proprioception, and the goal is updated online as each target is achieved, enabling multi-step object manipulation (see Appendix for details).

\textbf{Collecting Synthetic Image-Action Data.} We train a robot policy using Proximal Policy Optimization (PPO)~\cite{Schulman2017ProximalPO} to optimize our object-centric reward $r_{\rm obj}$. This policy learns to predict actions that enable the robot to manipulate objects matching the human demonstration. To learn robust behaviors, we randomize the starting pose of objects during RL training. The RL policy conditions on privileged simulation states $\pi_{\rm RL}(a | s)$ to output robot actions. After training the RL policy, we roll it out in our simulator to generate synthetic data of image-action pairs, only keeping successful trajectories. We systematically vary the simulation conditions by randomizing object starting positions, camera viewpoints, and lighting conditions across different rollouts. Each robot rollout is defined as $\xi_R = \{(o_R^t, a_R^t)\}_{t=1}^N$, an $N$ step trajectory of RGB image $o_R^t \in \mathbb{R}^{H \times W \times 3}$ and action $a_R^t$ pairs. This process builds a diverse synthetic dataset $D_{\rm synthetic}$ suitable for image-conditioned policy training.

\subsection{Sim-to-Real Transfer of Image-Based Policies}
\label{sec:sim2real}

We distill robot behaviors into image-conditioned policies trained on synthetic image-action pairs generated in simulation. To improve real-world transfer, we introduce an online domain adaptation technique that collects real image observations from closed-loop rollouts and automatically pairs them with simulated views of the same robot trajectories, then used to minimize the sim-to-real visual gap (Fig.~\ref{fig:approach-sim2real}, Right). Notably, this procedure does not require any robot teleoperation data.

\textbf{Training Image-Conditioned Policies.} 
Given the synthetic dataset $D_{\rm synthetic}$, we train an image-conditioned policy that operates directly on RGB observations without access to privileged state. We employ Diffusion Policy (DP)~\cite{chi2023diffusion}, a state-of-the-art behavior cloning architecture, to predict actions given the current image observation. The policy $\pi_{\rm img}(a | o)$ takes the current RGB observation $o_R^t$ as input and predicts a sequence of actions $\mathbf{a_R} = \{a_R^h\}_{h=1}^H$ over a horizon $H$ to complete the task.

\textbf{Auto-Calibration of Real and Sim.} While our photo-realistic simulation enables zero-shot transfer to the real world, visual discrepancies between simulated and real observations can still limit performance. To address this, we introduce an online domain adaptation technique that allows our policy's observation encoder to focus on task-relevant features by aligning images across domains. After deploying our initial policy in the real world, we collect image observations from these robot rollouts, \textbf{even including failures}. We then spawn our simulation with same initial state as the rollouts (using FoundationPose on the first frame), and replay the exact robot trajectories in simulation to create a paired dataset of real and simulated images (Fig~\ref{fig:approach-sim2real}, Right). The paired images can be used to supervise the policy's observation encoder by encouraging image pairs to map to the same embeddings while being distinguishable from images corresponding to different environment states.

This yields a dataset of paired observations $D_{\rm paired} = \{(o_R^{\rm sim}, o_R^{\rm real})\}_{i=1}^D$. During training, we supervise the policy with a standard behavior cloning loss on $D_{\rm synthetic}$ and apply an additional contrastive InfoNCE loss~\cite{Oord2018RepresentationLW} on $D_{\rm paired}$:

\begin{equation}
    \mathcal{L}_{\rm calibration} = \mathop{-\sum}\limits_{(o_R^{\rm sim}, o_R^{\rm real}) \in D_{\rm paired}} \frac{\exp(s(\phi(o_R^{\rm sim}), \phi(o_R^{\rm real}))/\tau)}{\sum\limits_{(\_, o_R^{'\rm real}) \in D_{\rm paired}} \exp(s(\phi(o_R^{\rm sim}), \phi(o_R^{'\rm real}))/\tau)}
\end{equation}

Here, $\phi$ is the policy’s image encoder, $s$ is cosine similarity, and $\tau$ is a temperature hyperparameter. This loss pulls together embeddings of corresponding simulation and real images, while pushing apart mismatched ones—guiding the encoder to focus on task-relevant semantics and reducing overfitting to simulation-specific features.

\section{Experiments \label{sec:experiments}}

\begin{figure}  
  \centering
  \includegraphics[width=\textwidth]{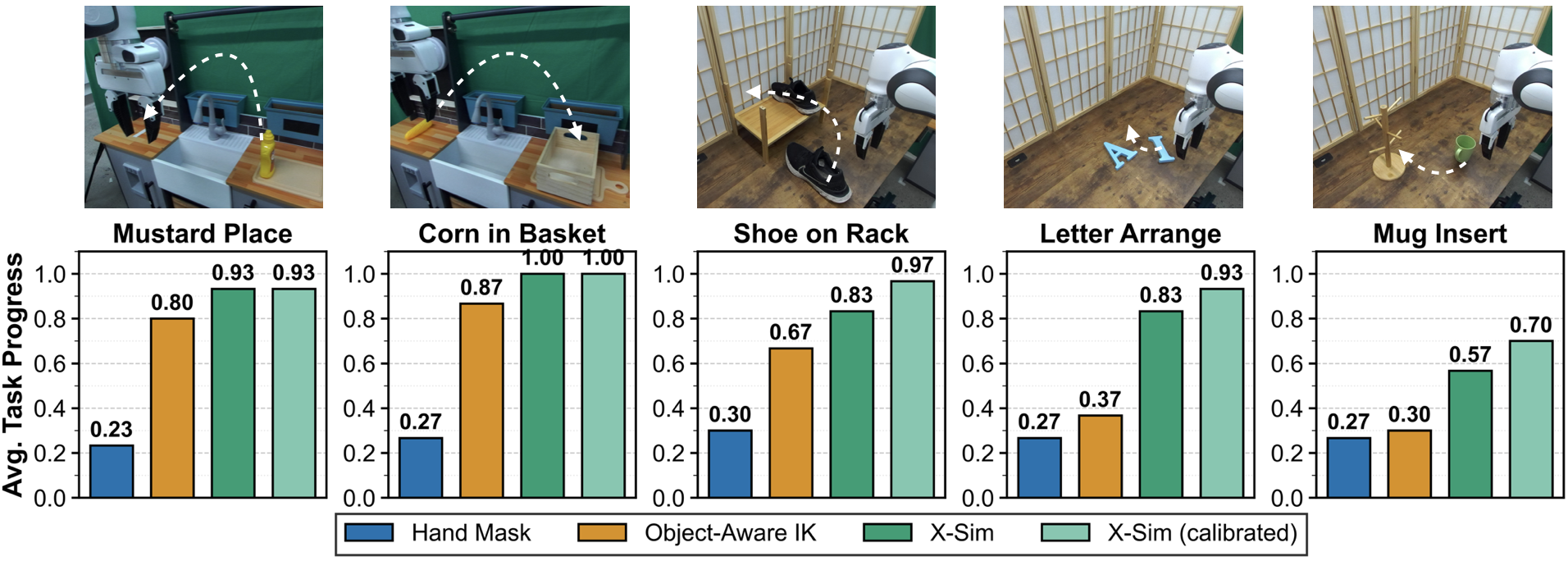}
  \caption{\textbf{Performance on Real-World Tasks:} We report \textit{Avg. Task Progress} on 5 tasks across two environments, and find that \methodname\ both with and without calibration consistently outperforms hand-tracking baselines that attempt to retarget human hand motion. A rough sketch of each task is visualized on top.}
  \label{fig:barplot}
\end{figure}

We evaluate whether \methodname\ can generate synthetic data from action-less human videos that is sufficient to train high-performing real-world robot policies. Our experiments span 5 tasks across two environments and aim to answer the following core questions:

\begin{enumerate}[leftmargin=*]
\itemsep0em 
    \item \textbf{Bridging the Embodiment Gap via Simulation:} How does \methodname\ perform across a variety of tasks compared to hand-tracking baselines when given a single human demonstration video? 
    \item \textbf{Sim-to-Real Policy Transfer:} What is the practicality of \methodname's sim-to-real transfer of image-based policies versus alternate observation representations?
    \item \textbf{Data Efficiency:} How does \methodname's performance scale with time spent on data collection compared to behavior cloning methods with teleoperated robot data?
    \item \textbf{Robustness to Test-Time Changes: } In what ways can \methodname\ generate synthetic data to enable real-world policy robustness beyond standard data collection procedures? 
\end{enumerate}

\textbf{Experimental Setup.} We conduct all experiments using a 7-DOF Franka arm across two real environments: \textit{Kitchen} and \textit{Tabletop} (Fig.~\ref{fig:barplot}). RGBD human videos are recorded using a ZED 2 stereo camera, with no constraints on motion or grasp style allowing for natural human execution. Tasks include pick-and-place (\texttt{Mustard Place}, \texttt{Corn in Basket}, \texttt{Shoe on Rack}), non-prehensile manipulation (\texttt{Letter Arrange}), and precise insertion (\texttt{Mug Insert}). We transfer human videos into simulation using our real-to-sim pipeline. For each task, we train privileged-state policies using PPO~\cite{Schulman2017ProximalPO} in ManiSkill~\cite{Mu2021ManiSkillGM} and randomize object and robot poses around the initial demonstration state. Then, the RL policy is distilled into an image-only Diffusion Policy~\cite{chi2023diffusion}. We assume approximate knowledge of the test-time camera viewpoint and render randomized viewpoints around it during training, adding robustness to small variations. At inference time, \methodname\ operates solely on real RGB image input. To align the observation encoder for \methodname\ (\textsc{Calibrated)}, we rollout 10 trajectories of \methodname\ in the real-world to collect paired real and sim data. More details about each task and hyperparameters are in the Appendix.

\textbf{Evaluation Metrics.} We report \textit{Average Task Progress} as our primary metric, which captures partial credit across distinct stages of task completion rather than relying on binary success. For grasp-based tasks (\texttt{Mustard Place}, \texttt{Corn in Basket}, \texttt{Shoe on Rack}, \texttt{Mug Insert}), progress is divided into three stages: approaching the correct object, successfully grasping it, and completing the manipulation to match the goal configuration from the human video. For the non-prehensile task (\texttt{Letter Arrange}), the stages correspond to approaching, rotating, and placing the object correctly. We evaluate all methods over 10 trials, each with slight variations in the object's initial position relative to the demonstrated human video. 

\begin{wrapfigure}{r}{0.5\textwidth}
    \vspace{-7mm}
    \centering
    \includegraphics[width=\linewidth]{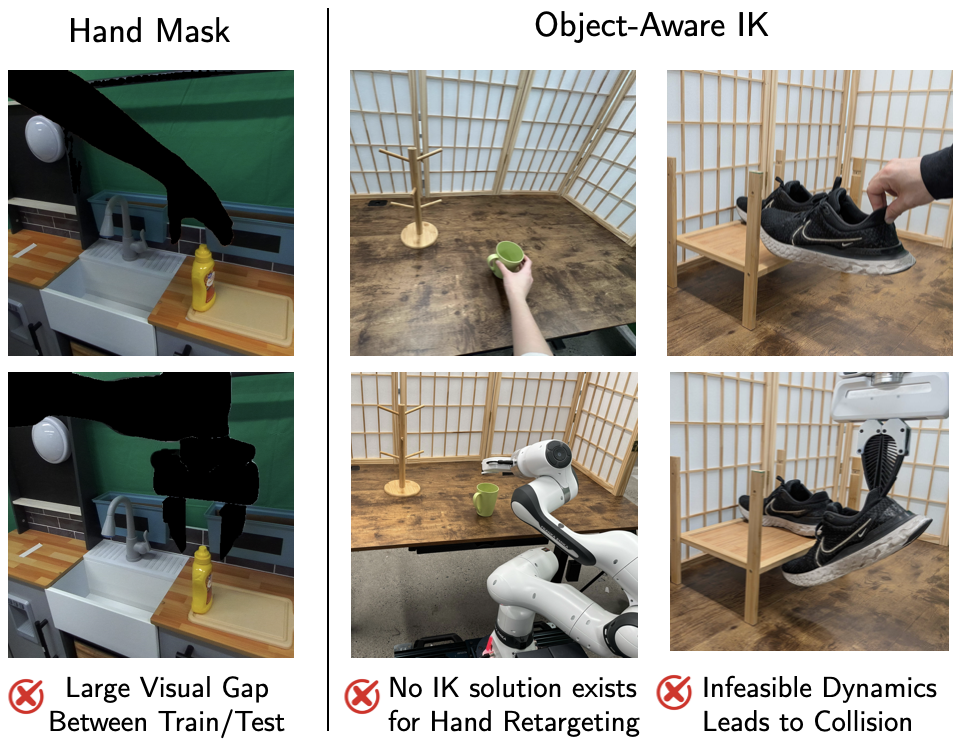}
    \vspace{-7mm}
    \captionsetup{width=0.5\textwidth}
    \caption{\textbf{Hand Re-targeting Failure Modes:} \textbf{Hand Mask} fails due to a significant visual domain gap between human and robots, even when the motions are physically feasible for the robot. \textbf{Object-Aware IK} fails under execution mismatch, as certain human hand motions are kinematically or dynamically infeasible.} \vspace{-6mm}
    \label{fig:retargeting}
\end{wrapfigure}

\subsection{Bridging the Embodiment Gap via Simulation}
\vspace{-1mm}
We evaluate whether \methodname\ can overcome the limitations of hand-retargeting approaches. We compare against two representative baselines:
\vspace{-3mm}
\begin{itemize}[leftmargin=*]
\itemsep0em 
    \item \textbf{Hand Mask:}~\cite{Lepert2025SHADOWLS, Lepert2025PhantomTR} Applies a black mask over the human hand in demonstration videos to train an image-conditioned behavior cloning policy. At inference time, the robot arm is similarly masked. This approach, used in PHANTOM~\cite{Lepert2025PhantomTR}, assumes all human hand poses can be replicated by the robot. Without this assumption, we do not overlay a robot arm during training.
    \item \textbf{Object-Aware Inverse Kinematics (IK):}~\cite{Vitiello2023OneShotIL, Li2024OKAMITH, Oord2018RepresentationLW} Extracts hand trajectories relative to nearby objects, and replays them by applying IK to move the robot end-effector along the same path.
\end{itemize}
\vspace{-3mm}
Neither baseline uses simulation. Both extract action labels from human hand pose estimates using HAMER~\cite{Pavlakos2023ReconstructingHI}, using the same procedure as PHANTOM~\cite{Lepert2025PhantomTR}. We evaluate \methodname\ and baselines across 10 real-world rollouts per task (Fig.~\ref{fig:barplot}). \textbf{Hand Mask} fails due to a large visual gap between human and robot observations, retaining only object location information and rarely progressing beyond the approach phase (Fig.~\ref{fig:retargeting}). \textbf{Object-Aware IK} performs well in \textit{Kitchen} tasks where human and robot have similar execution styles, but breaks down in \textit{Tabletop} tasks due to kinematic infeasibility and mismatched dynamics when directly mimicking human motions. In contrast, \textbf{\methodname}, even without sim-to-real calibration, learns feasible strategies in simulation and transfers them effectively—achieving consistently higher task progress and over 30\% gains in the most mismatched settings.

\vspace{-2mm}

\subsection{Sim-to-Real Policy Transfer}
\begin{wrapfigure}{r!}{0.35\textwidth}
    \vspace{-4mm}
    \centering
    \resizebox{0.35\textwidth}{!}{
    \begin{tabular}{c|c|c}
        \toprule
        \textbf{Metric $\downarrow$} & \textbf{\textsc{H2S2R}} & \textbf{\textsc{X-Sim}} \\
        \midrule
        Avg. Task Progress & 43.3\% & 83.3\%\\
        \bottomrule
    \end{tabular}
    }
    \vspace{-2mm}
    \caption{\small{We evaluate Avg. Task Progress of \methodname\ with image observations against a sim-to-real baseline that uses object state observations on the \texttt{Letter Arrange} task.}}
    \vspace{-4mm}
    \label{tab:sim2real_table}
\end{wrapfigure}

\vspace{-2mm}

\textbf{Comparison with State-Based Policy.} We evaluate \methodname's ability to transfer from simulation to the real world using only RGB images, and compare it to policy learning approaches based on privileged state, such as object poses. A closely related method, \textbf{Human2Sim2Robot}~\cite{Ga2025CrossingTH}, learns in simulation using accurate 6D object poses and attempts to replicate this setup in the real world through object tracking. However, even small tracking errors at inference can push pose-based policies out-of-distribution, leading to failure.
\begin{wrapfigure}[15]{r}{0.35\textwidth}
    \vspace{-3mm}
    \centering
    \includegraphics[width=\linewidth]{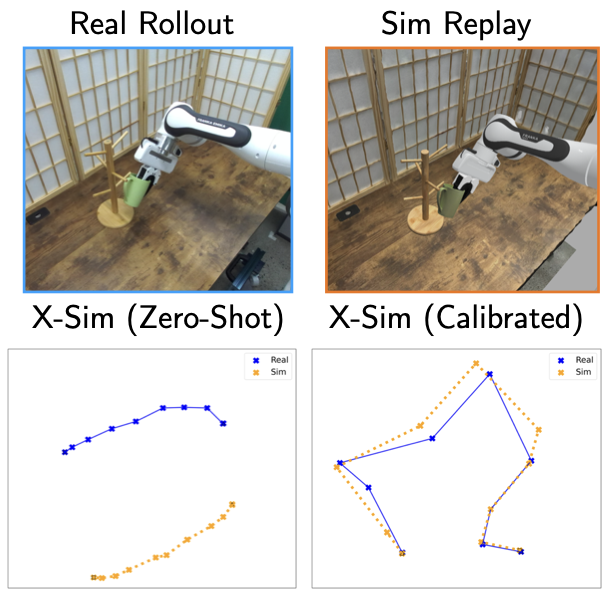}
    \vspace{-7mm}
    \captionsetup{width=0.35\textwidth}
    \caption{\textbf{Sim-to-Real Calibration:} We compare \methodname\ image embeddings using t-SNE before and after calibration for one rollout.
    }  
    \label{fig:tsne}
    \vspace{-6mm}
\end{wrapfigure}

\vspace{-2mm}
These methods often rely on precise observations that are hard to obtain in practice due to occlusions, depth noise, and imperfect vision models. In contrast, \textbf{\methodname}\ uses raw images, which provide a more robust and transferable representation. Image-based inputs are less sensitive to real-world noise and align well with modern visuomotor policy architectures. On the \texttt{Letter Arrange} task, \methodname\ significantly outperforms pose-based baselines in sim-to-real transfer (Table~\ref{tab:sim2real_table}), showing that images are a more practical and effective observation modality for real-world deployment.

\textbf{Calibration after Deployment.} Recent sim-to-real methods~\cite{Ankile2024FromIT, Nvidia2025GR00TNA} often rely on co-training with real-world demonstrations to bridge the domain gap. In contrast, \methodname\ uses only simulation data collected in a photorealistic environment, avoiding the need for teleoperation. While this reduces the observation gap, some visual discrepancies remain due to imperfections in 3D reconstruction and rendering. To address this, \methodname\ (\textsc{Calibrated}) aligns real and simulated observations online using closed-loop rollouts, as described in Sec.~\ref{sec:sim2real}.
Notably, this procedure is agnostic to success/failure, and can even benefit from unsuccessful rollouts. We find that \methodname\ (\textsc{Calibrated)} leads to additional benefits over our base method, with an average increase of 8\% in task progress across all tasks and most notably a 13\% increase for the most challenging task \texttt{Mug Insert}, indicating the ability to learn even from failures (Fig.~\ref{fig:barplot}). To further analyze the effects of our calibration procedure, we probe policy observation encoders on a paired simulation/real robot videos and plot the t-SNE embeddings over time in Fig.~\ref{fig:tsne}. \methodname\ (\textsc{Calibrated)} better aligns image embeddings compared to \methodname, ensuring that the policy avoids overfitting to domain-specific attributes with its calibration loss while still encoding task relevant features with its action prediction loss.

\vspace{-2mm}
\subsection{Data Efficiency}
\label{sec:experiments-data}
\vspace{-1mm}
\begin{wrapfigure}{r}{0.4\textwidth}
    \vspace{-2mm}
    \centering
    \includegraphics[width=\linewidth]{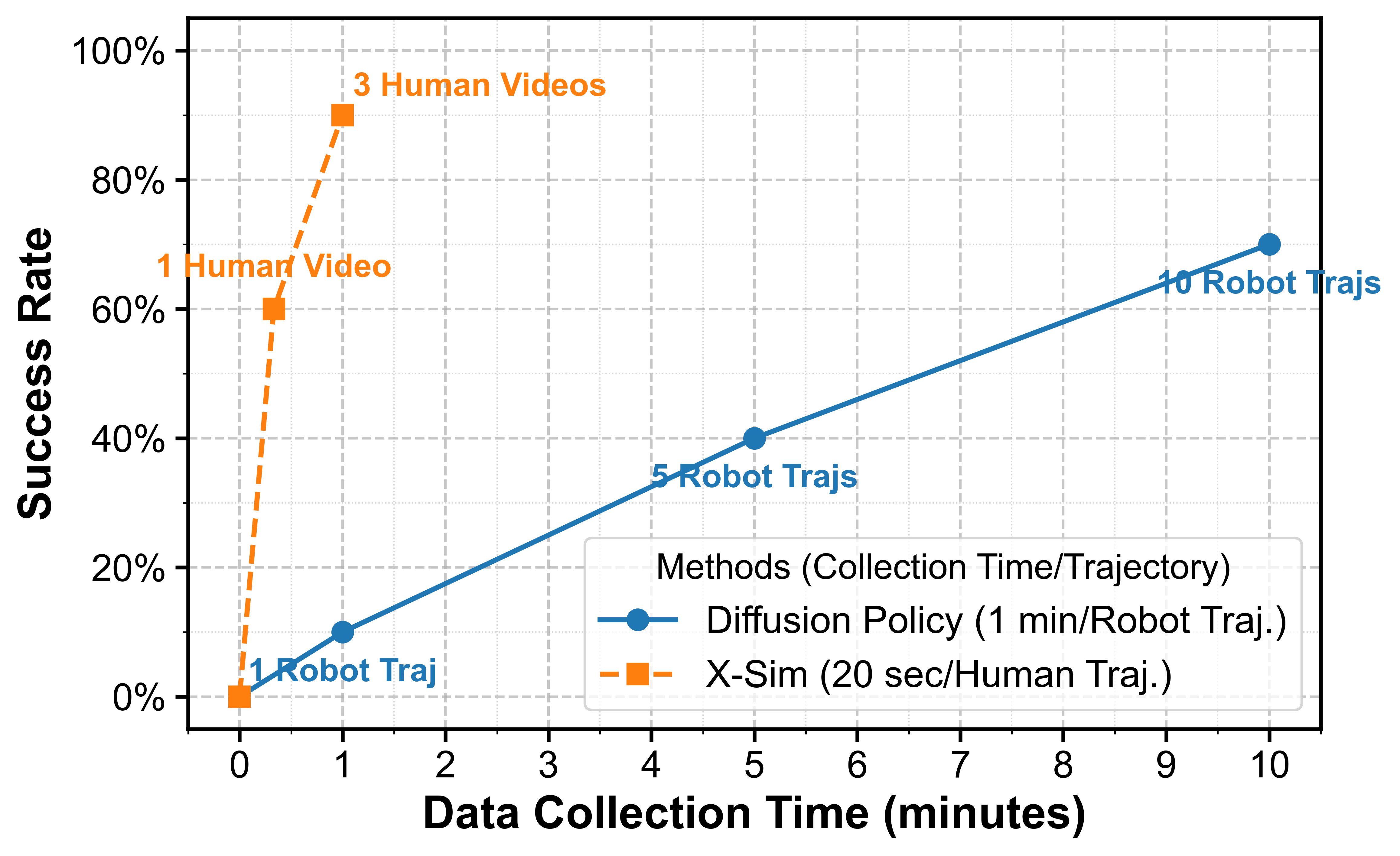}
    \captionsetup{width=0.4\textwidth}
    \caption{\textbf{Data Efficiency:} \methodname\ scales more efficiently with data collection time than behavior cloning from robot teleoperation, achieving comparable success on \texttt{Mustard Place} with 10$\times$ less time.
    }  
    \label{fig:scaling}
    \vspace{-6mm}
\end{wrapfigure}

We study how \methodname's performance scales with data by modifying the \texttt{Mustard Place} task to significantly broaden the initial state distribution of the mustard bottle (visualizations in the Appendix). In this setting, behavior cloning requires extensive robot teleoperation data to cover the distribution. In contrast, \methodname\ scales by collecting more human videos—which are faster to obtain (20s per video vs. 60s per robot demo)—and perturbing object poses in simulation for broader coverage. As shown in Fig.~\ref{fig:scaling}, \methodname\ achieves 90\% success with just 1 minute of human video data, compared to 70\% success with 10 minutes of robot demonstrations. This highlights \methodname's efficiency and scalability for training robust robot policies.
%


\vspace{-2mm}
\subsection{Robustness to Test-Time Changes}
\label{sec:experiments-viewpoint}
\vspace{-2mm}
\begin{wrapfigure}{r}{0.4\textwidth}
    \vspace{-4mm}
    \centering
    \resizebox{0.4\textwidth}{!}{
    \begin{tabular}{c|ccc}
    \toprule
    Train $\xrightarrow{}$  & \multicolumn{1}{c}{\multirow{2}{*}{Side}} & \multicolumn{1}{c}{\multirow{2}{*}{Frontal}} & \multicolumn{1}{c}{\multirow{2}{*}{Side \& Frontal}} \\  
    Test $\downarrow$ &  &  &  \\
    \midrule
      Side & 83.3\%  & 23.3\%  & \textbf{96.7\%} \\
     Frontal & 23.3\%  & 76.7\%  & \textbf{80.0\%} \\
     Novel & 33.3\%  & 30.0\%  & \textbf{53.5\%} \\
    \bottomrule
    \end{tabular}
    }
    \caption{\small{We show that we can flexibly collect image-action data in simulation from multiple viewpoints (Side and Frontal) with \methodname\ and train robust policies that generalize to novel viewpoints (\texttt{Shoe on Rack}).}}
    \vspace{-4mm}
    \label{tab:viewpoint_table}
\end{wrapfigure}
Image-conditioned policies are particularly sensitive to viewpoint bias, demanding additional data for each perspective. \methodname\ overcomes this by leveraging simulation to render trajectories from any desired view, enabling efficient coverage. We evaluate this by collecting simulated rollouts from \textit{Side} and \textit{Frontal} camera views, and training policies with data from each view individually and jointly for \texttt{Shoe on Rack}. As shown in Fig.~\ref{tab:viewpoint_table}, combining diverse viewpoints in simulation significantly improves generalization, even to unseen camera angles. More details are in the appendix.

\vspace{-3mm}
\section{Discussion}
\methodname\ presents a scalable framework for learning robot manipulation policies from human videos without requiring action labels or robot teleoperation. By leveraging object motion as a dense supervisory signal and training in photorealistic simulation, \methodname\ bridges the embodiment gap and transfers image-conditioned policies to the real-world using online domain adaptation. Across five manipulation tasks, it improves task progress on average by over 30\% compared to hand-tracking baselines, matches behavior cloning performance with 10× less data, and generalizes to novel viewpoints and test-time changes. While this work focuses on training policies from scratch, the same real-to-sim-to-real approach can naturally extend to fine-tuning pre-trained robot learning models, including foundation models such as PI-0.5~\cite{Intelligence2025pi} and OpenVLA~\cite{Kim2024OpenVLAAO}. By generating targeted synthetic rollouts conditioned on human video demonstrations, \methodname\ could adapt generalist policies to new tasks and embodiments in a low-cost, scalable manner and offers a complementary path toward efficient specialization of large robot models without requiring new robot data.

\section{Limitations}

In this paper, we chose to maximize the ability of the real-to-sim pipeline by making simplifying assumptions, while still maintaining the input/output contract (images to actions) that is most practical to deploy in unstructured environments. This is because the focus of the paper is to show the effectiveness of image-based policy transfer given ideal real-to-sim transfer. However, we acknowledge that while \methodname\ provides an effective approach for learning robot policies from human videos, its application to unstructured, in-the-wild internet videos remains an open challenge. Below, we outline key assumptions that limit \methodname's current ability to move towards this broader vision and suggest pathways towards their solutions in the near future:

\textbf{Requiring Object Meshes for Tracking.}
Our pipeline currently uses FoundationPose, which requires a 3D object mesh for tracking, limiting applicability to videos where we either don't know or don't have the object mesh manipulated by the human. One way to extend this to internet videos is by estimating approximate meshes directly from using tools like InstantMesh~\cite{Xu2024InstantMeshE3}. Alternatively, object meshes can be retrieved from large 3D asset libraries~\cite{Li2022BEHAVIOR1KAB}, as shown in prior work on digital cousin generation~\cite{Dai2024AutomatedCO}, which suffices since simulation is only used for synthetic data generation.

\textbf{Restricted to Rigid Object Manipulation.}
Our current pipeline relies on tracking object states through 6D poses, which limits it to rigid objects and excludes articulated or deformable items commonly seen in real-world tasks. For articulated objects like drawers or doors, recent vision research~\cite{Xia2025DRAWERDR} has shown that visual priors and foundation models can be used to identify and track articulation parameters from RGB input. For deformable objects, emerging representations like particle-based models~\cite{AbouChakra2024PhysicallyEG} offer promising avenues for capturing non-rigid dynamics. While these approaches are still maturing, our framework can continue to improve rigid manipulation skills, and its image-conditioned policies may complement existing models trained on separate data to handle deformables and articulations more effectively.

\textbf{Environment Scan for Generating Simulation.}
\methodname\ currently requires an explicit 3D scan of the environment to reconstruct the simulation scene, which limits its applicability to scenarios where such scans are unavailable. Recent works like St4RTrack~\cite{Feng2025St4RTrackS4} have shown that it is possible to generate both geometric and visual reconstructions directly from monocular human videos. While these methods typically rely on dynamic camera motion, many human video datasets—such as Ego4D~\cite{Grauman2021Ego4DAT}—naturally satisfy this condition, offering a viable path toward removing the explicit environment scanning requirement.

\textbf{Estimating Physics Parameters from Vision Alone.}
In this work, we use default physics parameters—such as mass, friction, and stiffness—for simulation, rather than estimating them from the human video. However, recent approaches suggest viable paths forward: vision language models (e.g., GPT) can provide plausible physics guesses given object categories or visual context~\cite{xia2025drawer}, and domain randomization can be applied around these estimates to build robustness. Additionally, while our proposed online sim-to-real calibration targets visual alignment, the same framework could be extended to iteratively adapt physical parameters by comparing real and simulated rollouts—enabling self-supervised refinement of both perception and dynamics.


\section{Acknowledgements}
Sanjiban Choudhury is supported in part by Google Faculty Research Award, OpenAI SuperAlignment Grant, ONR Young Investigator Award, NSF RI \#2312956, and NSF FRR \#2327973. Wei-Chiu Ma is supported in part by a gift from Ai2, a NVIDIA Academic Grant, and DARPA TIAMAT program No. HR00112490422.

\clearpage


\bibliography{refs}  
\clearpage
\section{Appendix}
\subsection{Task Descriptions}
\begin{figure}[h]
  \centering
  \includegraphics[width=\textwidth]{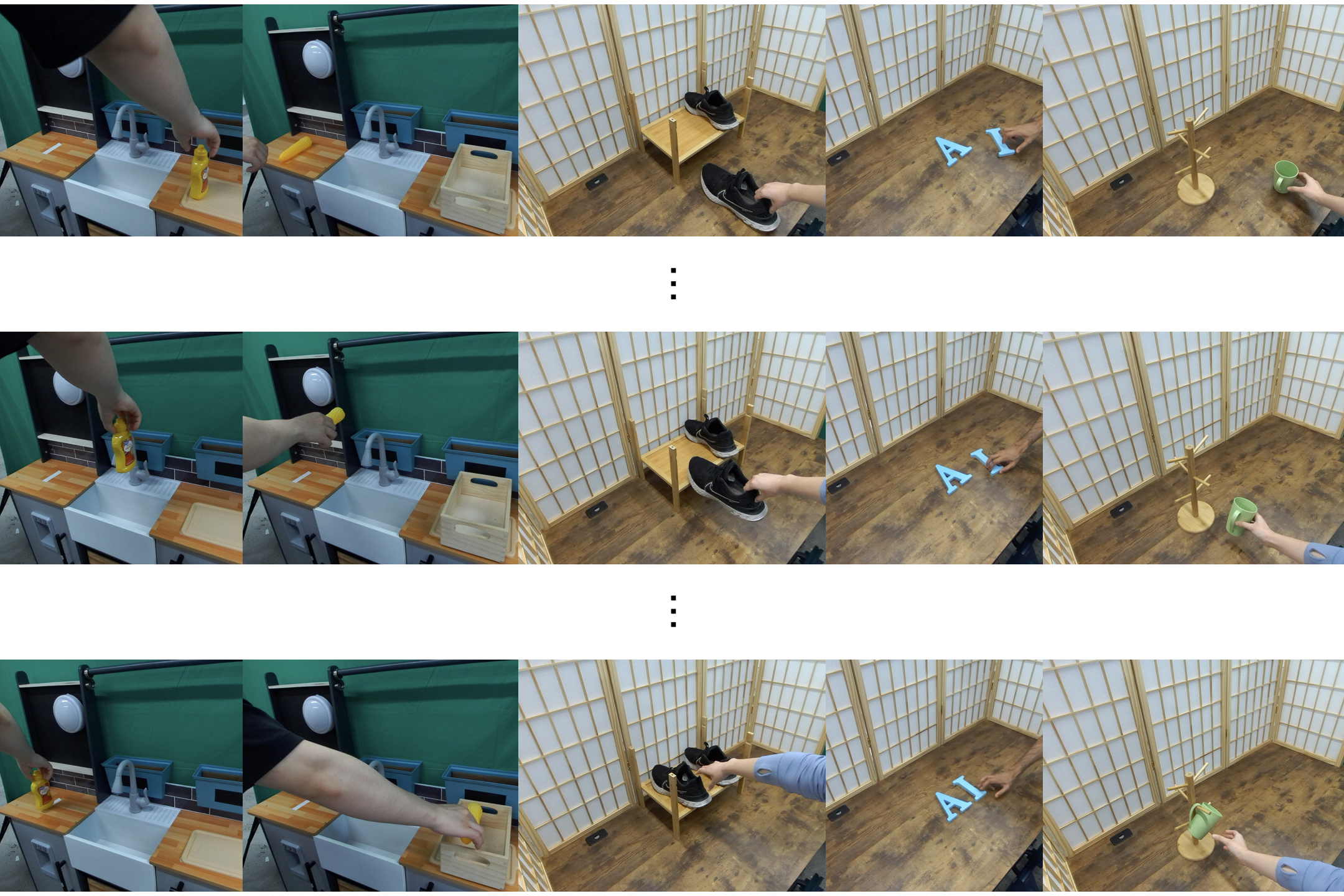}
  \caption{Visualization of tasks that we report results for in Fig.~\ref{fig:barplot}}
  \label{fig:task-imgs}
\end{figure}

We provide descriptions and visualizations (Fig.~\ref{fig:task-imgs}) of tasks we report results for in Fig.~\ref{fig:barplot}.
\begin{itemize}
    \item \texttt{Mustard Place}: Pick up the Mustard bottle from the right side of the kitchen countertop and place it on the left side.
    \item \texttt{Corn in Basket}: Pick up the corn from the left side of the kitchen countertop and put it inside of the basket.
    \item \texttt{Shoe on Rack}: Pick up the left shoe and place it on top of the shoe rack, next to the right shoe.
    \item \texttt{Letter Arrange}: Move the letter \textit{'I'} next to the letter \textit{'A'} so that they are aligned.
    \item \texttt{Mug Insert}: Insert the mug's handle onto the holder.
\end{itemize}

\subsection{Real-to-Sim Scans}
In order to transfer our environments and objects into simulation, we employ 2D Gaussian Splatting~\cite{Huang20242DGS}. We take videos (multi-view images) of the environment for $<2$ minutes, which are supplied as input to the module to get a photo-realistic 3D reconstruction of the scene. Individual objects to be tracked are scanned with Polycam~\cite{polycam2020}, a phone app, with a similar procedure in $<1$ minute per object. The environment and objects are scaled manually to the correct size before being transferred into simulation, though alternate calibration methods could be used to automate this process.

\clearpage
\subsection{RL Training Details}
\subsubsection{PPO Hyperparameters}
We provide details of hyperparameters (Table~\ref{tab:ppo_hyperparams}) used for training privileged-state PPO~\cite{Schulman2017ProximalPO} policies in simulation. 
\begin{table}[h]
\centering
\caption{PPO Hyperparameters}
\begin{tabular}{lc}
\toprule
\textbf{Hyperparameter} & \textbf{Value} \\
\midrule
Learning rate & $3 \times 10^{-4}$ \\
Discount factor ($\gamma$) & 0.8 \\
GAE parameter ($\lambda$) & 0.9 \\
Clipping parameter ($\epsilon$) & 0.2 \\
Value function coefficient & 0.5 \\
Entropy coefficient & 0.0 \\
Target KL divergence & 0.1 \\
Maximum gradient norm & 0.5 \\
Minibatch size & 9,600 \\
Number of parallel environments & 1,024 \\
\midrule
Actor network & MLP (state dim $\rightarrow$ 256 $\rightarrow$ 256 $\rightarrow$ 256 $\rightarrow$ action dim) \\
Critic network & MLP (state dim $\rightarrow$ 256 $\rightarrow$ 256 $\rightarrow$ 256 $\rightarrow$ 1) \\
Activation function & Tanh \\
\midrule
Optimizer & Adam \\
Adam epsilon & $1 \times 10^{-5}$ \\
\bottomrule
\end{tabular}
\label{tab:ppo_hyperparams}
\end{table}

\subsubsection{Simulation State Space}
The state-based observation space in simulation consists of the following components: 
\begin{table}[h]
\centering
\caption{Observation Space Components}
\begin{tabular}{ll}
\toprule
\textbf{Component} & \textbf{Description} \\
\midrule
\texttt{ee\_pose} & End-effector pose (position and orientation) \\
\texttt{gripper\_width} & Gripper opening width \\
\texttt{achieved\_goal} & Current object poses\\
\texttt{desired\_goal} & Target waypoint poses for objects \\
\texttt{goal\_position\_diff} & Position difference between current and target poses \\
\texttt{goal\_rotation\_diff} & Angular difference between current and target orientations \\
\texttt{is\_grasped} & Binary object grasp status (if applicable)\\
\bottomrule
\end{tabular}
\label{tab:observation_space}
\end{table}

\subsubsection{Reward Function Formulation}
We provide the complete object-centric reward function implementation proposed in Sec.~\ref{sec:approach}:

\paragraph{Approach Reward.} The $r_{\rm approach}$ component encourages the agent to approach the target object with:
\begin{equation}
r_{\rm approach} = \left(1 - \tanh(kd_{\rm obj})\right)
\end{equation}
\noindent where $d_{\rm obj}$ is the distance between the end-effector and the current target object and $k$ is a constant scaling factor. 

\paragraph{Goal Reward.} $r_{\rm goal}$ penalizes positional and rotational deviations from the target state:
\begin{equation}
r_{\rm goal} = \left(1 - \tanh(\alpha_d \cdot d_{\rm pos}(s^B_H, s^t_R))\right) + \left(1 - \tanh(\alpha_{\theta} \cdot d_{\rm rot}(s^B_H, s^t_R))\right) + 2i_{\rm waypoint}
\end{equation}
\noindent where $d_{\rm pos}(\cdot)$ measures the Euclidean distance, and $d_{\rm rot}(\cdot)$ computes the quaternion angular difference, $\alpha_d$ and $\alpha_{\theta}$ are scaling factors for each waypoint automatically computed from the demonstration, and $i_{\rm waypoint}$ is the current waypoint index to serve as a bonus for progressing through the task. Note that the \texttt{desired\_goal} in the observation is updated when the current goal is reached within an $\epsilon$ threshold, and in practice we sample $N$ object waypoints from the human video to summarize the demonstration.

The goal reward also has additional terms: $r_{\rm static}$ encourages stability of the robot when objects are correctly positioned, $r_{\rm success}$ provides a $+1$ bonus upon task completion (objects are placed in their goal configuration), and $r_{\rm grasp}$ is an optional binary reward to encourage grasps for non-prehensile tasks.

\paragraph{Complete Reward.}
The final reward is $r_{\rm obj} = r_{\rm approach} + r_{\rm goal}$.

\subsection{Image-Conditioned Policy Training Details}
\subsubsection{Synthetic Data Collection}
We provide details on randomization parameters (Table~\ref{tab:randomization_params}) used when collecting synthetic data for $D_{\rm synthetic}$ (Sec.~\ref{sec:approach}).

\begin{table}[h]
\centering
\caption{Environment Randomization Parameters}
\begin{tabular}{lc}
\toprule
\textbf{Parameter} & \textbf{Value} \\
\midrule
\multicolumn{2}{l}{\textit{Object Randomization}} \\
Initial pose position noise (XY) & $\pm 0.025$ m \\
Initial pose rotation noise & $\pm \pi/8$ rad \\
\midrule
\multicolumn{2}{l}{\textit{Robot Randomization}} \\
Initial robot joint angle noise & $\pm 0.02$ rad \\
\midrule
\multicolumn{2}{l}{\textit{Camera and Lighting (Evaluation)}} \\
Camera position variation & $\pm 0.03$ m \\
Camera target position variation & $\pm 0.03$ m \\
Lighting configurations & 4 presets \\
\bottomrule
\end{tabular}
\label{tab:randomization_params}
\end{table}

For each task, we collect 500 visuomotor demonstrations in simulation by applying these randomization parameters.

\subsubsection{Image-Conditioned Diffusion Policy Training}
We provide hyperparameters (Table~\ref{tab:diffusion_policy_hyperparams}) for training Diffusion Policies~\cite{chi2023diffusion}, where input is simply an image of the current state and output is 7-dimensional delta actions in end-effector space (3 position actions, 3 rotation actions, 1 gripper action). 
\begin{table}[h]
\centering
\caption{Diffusion Policy Training Hyperparameters}
\begin{tabular}{lc}
\toprule
\textbf{Parameter} & \textbf{Value} \\
\midrule
Diffusion timesteps (training) & 100 \\
Diffusion timesteps (inference) & 10 \\
\midrule
Backbone CNN & ResNet18 \\
Image size & 960 $\times$ 720 $\rightarrow$ 96 $\times$ 96 \\
Image feature dimension & 512 \\
Diffusion step embedding dimension & 128 \\
Kernel size & 5 \\
Normalization layer & GNN \\
\midrule
Action horizon & 2 \\
Prediction horizon & 8 \\
Shift padding & 6 \\
\midrule
Batch size & 64 \\
Learning rate & $1 \times 10^{-4}$ \\
Weight decay & $1 \times 10^{-6}$ \\
Gradient clipping & 5.0 \\
EMA decay rate & 0.01 \\
\midrule
Action prediction loss weight & 1 \\
Auto-calibration loss weight & 0.1 (if applicable)\\
\bottomrule
\end{tabular}
\label{tab:diffusion_policy_hyperparams}
\end{table}

\subsection{Ablation Visualizations}

\subsubsection{Data Efficiency}
We provide visualizations (Fig.~\ref{fig:efficiency-viz}) of the initial state distribution of the Mustard bottle as training input for the data efficiency ablation in Sec.~\ref{sec:experiments-data}. The robot teleoperation data takes 10 minutes to collect, while the human videos take 1 minute to collect. In simulation, the starting poses of the object are perturbed to enable robustness of the RL policy and diversity in during synthetic data collection. The evaluation distribution is across the cutting board.

\begin{figure}[h]
  \centering
  \includegraphics[width=\textwidth]{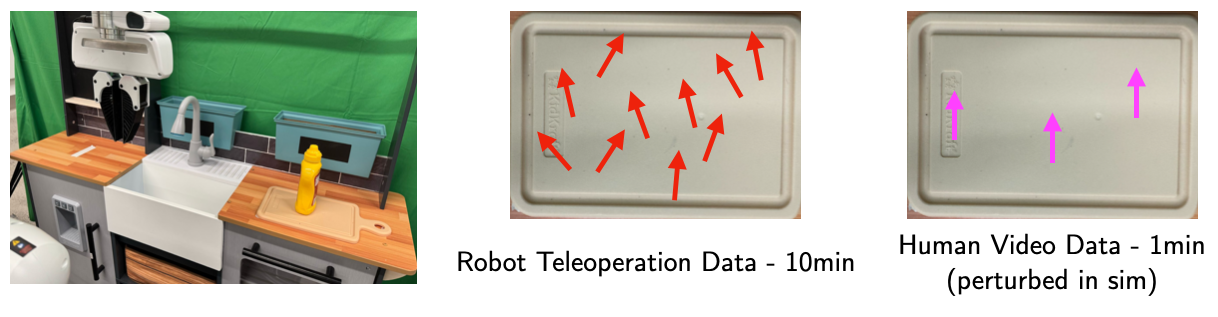}
  \caption{Visualization of training states for results in  Fig.~\ref{fig:scaling}}
  \label{fig:efficiency-viz}
\end{figure}

\subsubsection{Viewpoint Changes}
We provide visualizations (Fig.~\ref{fig:efficiency-viz}) of the three different viewpoints that we study at train/test time in Sec.~\ref{sec:experiments-viewpoint}.

\begin{figure}[h]
  \centering
  \includegraphics[width=\textwidth]{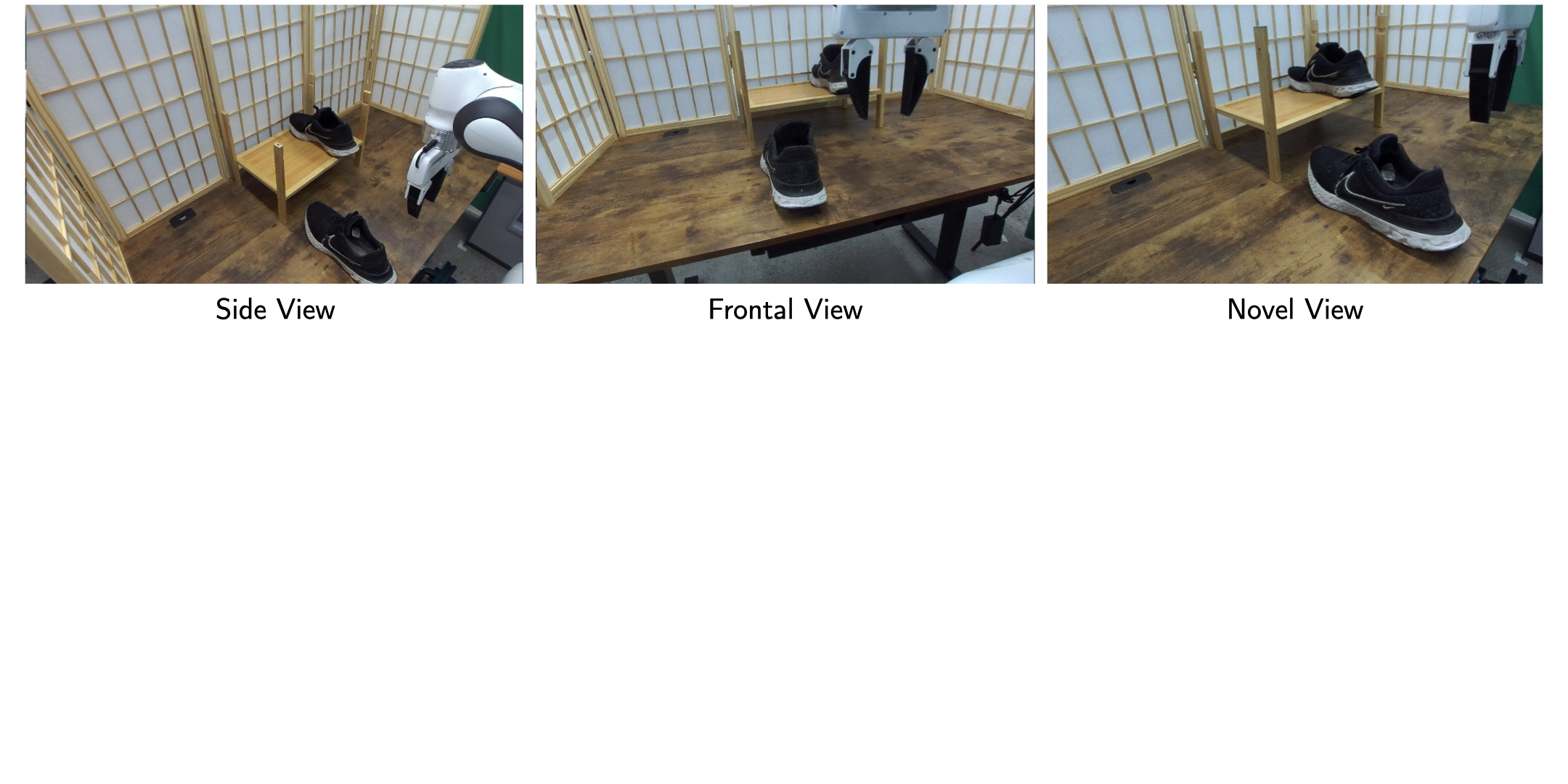}
  \vspace{-44mm}
  \caption{Visualization of viewpoints for results in  Fig.~\ref{tab:viewpoint_table}}
  \label{fig:efficiency-viz}
\end{figure}

\end{document}